\documentclass[letterpaper, 10 pt, conference]{ieeeconf}
\usepackage{newtxtext}

\IEEEoverridecommandlockouts

\usepackage{lipsum}
\usepackage{cite}
\usepackage{amsmath,amssymb,amsfonts}
\usepackage{algorithmic}
\usepackage{graphicx}
\usepackage{textcomp}
\usepackage{xcolor}
\usepackage{url}
\usepackage{pifont}
\usepackage{xcolor}
\usepackage[T1]{fontenc}
\def\BibTeX{{\rm B\kern-.05em{\sc i\kern-.025em b}\kern-.08em
    T\kern-.1667em\lower.7ex\hbox{E}\kern-.125emX}}

\newcommand{\skill}[1]{\texttt{#1}}
\newcommand{\cmark}{\textcolor{green!60!black}{\ding{51}}}
\newcommand{\xmark}{\textcolor{red}{\ding{55}}}

\begin{document}

\title{SkilledVLA: Integrating Classical Skills into \\Vision-Language-Action Models}

\author{Rumi Loghmani$^{1,2}$,
Mohamed Kari$^{2}$,
Dong-Ki Kim$^{2}$,
Y. Isabel Liu$^{2,3}$,
Shayegan Omidshafiei$^{2}$,
Abraham George$^{2}$
\thanks{$^{1}$Carnegie Mellon University, Pittsburgh, PA, USA.}%
\thanks{$^{2}$FieldAI, Irvine, CA, USA.}%
\thanks{$^{3}$Massachusetts Institute of Technology, Cambridge, MA, USA.}%
}

\maketitle

\begin{abstract}

Vision-language-action models (VLAs) are a powerful tool for robot control, presenting a unified architecture that can learn to solve a wide range of tasks and generalize to novel scenes and instructions without requiring human-engineered controllers. However, by embracing end-to-end learned control, VLAs discard decades of progress in classical robotics, along with the speed, reliability, and efficiency of classical solutions. In this work, we seek to combine the generalizability of VLAs with the performance of classical skills. We present SkilledVLA, a system that adds the ability to invoke classical skills directly into a VLA. SkilledVLA modifies a traditional VLA by adding two auxiliary outputs per arm: a \textit{classifier} head, which selects which skill (if any) to invoke, and a \textit{target} head to parameterize the classical skills. By directly incorporating skill use into the action-space of the VLA, we seamlessly integrate classical skills at the VLA's native inference rate. When evaluated on a real Unitree G1, SkilledVLA raises end-to-end success from 5\% to 60\% on a bimanual ballistic toss task and from 20\% to 50\% on a long-horizon toast-preparation task. Against a suite of six DexMimicGen simulation tasks across two embodiments, the same architecture outperforms a vanilla fine-tuned VLA, a hierarchical VLA, and a VLM orchestrator. Gains were most obvious in the tasks where tolerance was the tightest.
For demonstration videos and more information see: https://rumilog.github.io/skilledvla/
\end{abstract}

\section{Introduction}
Classical model-based controllers have long dominated the robotics discipline, because they have high precision, force control, verifiable guarantees, and high-speed rollouts that make them suitable for industrial deployment.
A major downside of these classical controller-based systems is that they are generally not generalizable.
If the scene or instructions for a task change, classical controllers become unfit and need re-engineering.
Vision-Language-Action models (VLAs) are end-to-end learned policies that directly predict robot actions from vision observations and language commands, enabling generalizability across objects, scenes, and instructions. However, they  lack the precision, reactivity, and consistency of classical solutions. For example, if the VLA uses action chunking, then the system's reaction speed is bounded by the chunk length.

\begin{figure}[t!]
\centering
\includegraphics[width=\columnwidth]{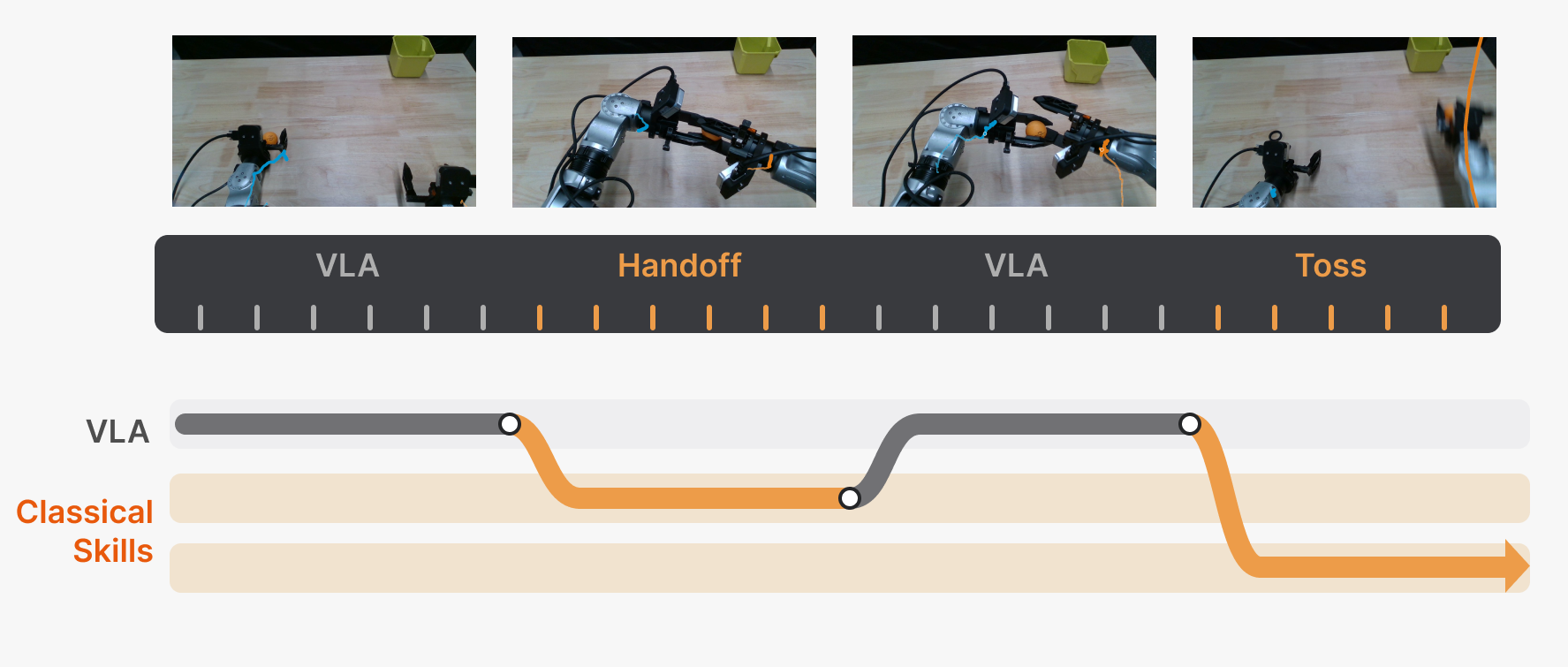}
\caption{SkilledVLA at inference on a single episode of the ball toss task. The \textit{classifier} head predicts a skill token at every timestep, determining whether robot actions are controlled by the finetuned VLA or a classical skill controller. When activated, classical skills are allowed to run until completion or failure, at which point control is reassigned based on the \textit{classifier's} prediction. The result is smooth robot actions smoothly interleaving VLA learned control and classical control}
\label{fig:intro}
\end{figure}

Prior works have tried to bridge the gap between classical control and VLAs through Vision-Language Models (VLMs) as orchestrators~\cite{ahn2022can,liang2023code,huang2023voxposer}, VLAs with high-level subtask conditioning, and hybrid action-space controllers.
VLM-as-orchestrator approaches provide a VLM with a predefined library of classical skills which the VLM can call to solve the task.
However, these systems struggle with the long inference times of many VLMs. Hierarchical VLAs like $\pi_{0.5}$~\cite{intelligence2025pi_} train both a high-level planner, which predicts the robot's high-level sub-task (often in language), and a low-level action expert, which generates robot actions conditioned on the sub-task.
This approach grants the system high-level reasoning, but by using only learned low-level controllers, it preserves many of the performance limitations of end-to-end learned methods.
Hybrid action-space controllers switch action representations at inference; previous papers such as HYDRA~\cite{belkhale2023hydra} and SPHINX~\cite{sundaresan2025s} switch control between a classical planner for large motions, and an imitation-learned policy for precision movements that require closed-loop feedback.
However, these works only use classical control for a single skill, large scale motion, and do not include feedback.

In this work, we present SkilledVLA, which tackles the challenge of combining VLAs with classical control by integrating classical skills into the VLA action space. To this end, SkilledVLA adds two additional outputs to the VLA: a \textit{classifier} head, which determines which controller should be used (either the VLA or one of the classical skills) and a \textit{target} head, which outputs parameterization information for the associated skill (e.g., the goal pose for the move-to skill). An example of a SkilledVLA rollout showing these skill activations at inference can be seen in Fig. \ref{fig:intro}.

By incorporating the ability to directly use a library of skills into a vision-language-action policy, SkilledVLA can reap the performance benefits of classical robotics while maintaining the generalizability of large end-to-end learned models. We validate SkilledVLA on a real Unitree G1 humanoid with Dex1 grippers, showing that by adding classical skills to a VLA, the model's success rate increased from 5\% to 60\% on a bimanual pick--handoff--toss task and from 20\% to 50\% on a toast-preparation task, and outperformed both hierarchical VLAs and VLM-as-orchestrators. Additional experiments across six bimanual manipulation tasks and two embodiments in simulation further support our results. Our contributions are:

\begin{itemize}
  \item \textbf{Skill-integrated action space.} A hybrid architecture that integrates classical skills directly into a VLA's action space via an auxiliary classifier and target heads.
  \item \textbf{Native-rate skill dispatch.} Skill selection built into the VLA backbone, so handoff between learned control and classical skills runs at the policy's native inference rate.
  \item \textbf{Real-robot and simulation evaluation.} A comparison of a fine-tuned VLA without skills, a hierarchical VLA, a VLM-as-orchestrator, and SkilledVLA, both in sim and on hardware (Unitree G1 with Dex1 grippers).
\end{itemize}

\section{Related Work}

\subsection{Vision-Language-Action Models}
Vision-Language-Action models (VLAs), such as GR00T~\cite{bjorck2025gr00t}, $\pi_0$~\cite{black2024pi_0}, and OpenVLA~\cite{kim2024openvla}, leverage recent advances in internet-scale-trained vision-language models (VLMs) for robotic control.
These models add an action decoder to an existing VLM backbone to directly map from images and language observations to actions, typically using diffusion or flow matching for action prediction ~\cite{chi2025diffusion,lipman2022flow}. Hierarchical VLAs, such as $\pi_{0.5}$~\cite{intelligence2025pi_}, RT-H~\cite{belkhale2024rt}, and Hi~Robot~\cite{shi2025hi}, seek to combine the approaches by training a high-level planner that creates intermediate instructions from the textual prompt and conditions a lower-level executor. SkillVLA~\cite{zhai2026skillvla} similarly uses hierarchical language conditioning, with per-arm learned action generation to enable bimanual skill reuse. These executors, though, are still learned policies, inheriting many of the limitations of end-to-end VLAs, with the main upside of these systems lying in their ability to work on longer-horizon tasks. We differ from all three in that SkilledVLA adds classical skills onto an end-to-end VLA so that phases where a structured controller is available are executed with higher reliability, while also being able to fall back upon the VLA's general action knowledge.

\subsection{VLM-as-Orchestrator}

VLM-as-orchestrator approaches leverage off-the-shelf VLMs for robot control by providing access to tools or skill libraries. Given a task description and scene observations, the model selects and parameterizes skills to execute. Building on LLM-based planners such as SayCan~\cite{ahn2022can}, systems such as ViLa~\cite{hu2023look} and Pigey~\cite{galanti2026addressing} incorporate visual observations into skill selection. Skills can be invoked individually (similar to tool invocation \cite{yao2022react}) or through a VLM-generated script to allow for multi-skill chaining and conditional logic, as demonstrated by Code as Policies~\cite{liang2023code}. The underlying tools may expose classical controllers or learned policies, including pretrained VLAs. This approach enables reuse of existing skills and code without robot-specific fine-tuning of the orchestrating model. However, these systems are unable to plan or replan at high frequencies and are limited to the skills in their library~\cite{galanti2026addressing}. In contrast, our method predicts skill dispatch directly from the VLA's representations while retaining its learned action head, avoiding the additional inference latency of a separate orchestration model.

\subsection{Hybrid Action Spaces}

Hybrid action space models switch action representations between learned and deterministic movements. An example of such a model is HYDRA~\cite{belkhale2023hydra}, which pairs a mode classifier with a waypoint head and an action head; SPHINX~\cite{sundaresan2025s} switches between waypoints and a dense end-effector action head. These methods deterministically move to learned waypoints, but when contact-rich, precise actions are required, they switch back to learned policies. SkilledVLA extends this approach, adding a diverse library of additional skills beyond ``move-to'', along with control over skill handoff, skill parameterization, and skill termination, allowing for a much richer integration of low-level controllers. Additionally, our method allows for per-arm skill activation, allowing for complex bimanual behaviors.

\subsection{Classical Control with Learned Policies}
A wide range of existing strategies exist for solving manipulation tasks, which for simplicity we refer to as ``classical skills'', although we use this term to refer to any skill that is not end-to-end learned.
For this work, we are specifically interested in obstacle-avoidant path planning, tossing, grasping, and insertion. Obstacle-avoidant path planning seeks to find an optimal path through a cluttered scene without colliding with objects.
cuRobo\cite{sundaralingam2023curobo} does this by wrapping the robot in collision spheres that are checked against objects in the scene.
Penetration is turned into a collision cost that cuRobo's trajectory optimization drives to zero while solving for path trajectory.
The tossing skill presents an interesting challenge as the robot cannot control the object during flight.
TossingBot~\cite{zeng2020tossingbot} uses a ballistic solver to calculate a release position and velocity that will result in the object landing at a goal location.
For grasping, we take inspiration from the classical antipodal grasping line of work~\cite{nguyen1988constructing,ferrari1992planning,ten2017grasp}, where a system enumerates antipodal contact pairs on the object's point cloud, scores each with a grasp-quality metric, and executes the best feasible grasp, so a stable pinch is found from geometry alone rather than from a per-object rule. The insertion skill is based on peg-in-hole strategies, such as the one presented by \cite{chhatpar2001search} in which the peg is first aligned with the hole's axis and then inserted, retreating on jams and continuing with a spiral search and push to find the right orientation for insertion.

\section{Problem Formulation}

Our goal is to develop a controller ${f}$ that maps robot observations $O$ to robot actions $A$, $f:O\to A$. We assume that the robot is a bimanual manipulator which, at each timestep $t$, executes an action $a_t \in A$. Additionally, the robot's observation $o_t \in O$ consists of RGB-D images, a language instruction, and the robot's proprioceptive state.

Additionally, we assume that our system has access to a skill library $\mathcal{C}$, which contains $n$ classical skills, $C_i \in \mathcal{C}$ where each skill is a functional mapping from observations $O$ to desired actions $A$, conditioned on skill-specific target information $G_i$, $C_i: (O, G_i) \to A$.

We also assume that, in addition to our classical skills, our skill library $\mathcal{C}$ contains a VLA controller, $C_\text{VLA}$, which converts VLA outputs, $G_\text{VLA}$ (i.e., delta goal pose action chunks) into robot actions $A$, $C_\text{VLA}: (O, G_\text{VLA}) \to A$.

Given this formulation, we can break the policy mapping $f$ into two stages: first, skill selection, $S$, which maps the observations $O$ to skill selection $i$ and target information $G_i$, $S: O  \to (i, G_i)$ and second, the skill controller, $C_i$, which is chosen by the controller index and  uses the output target information to map the observations to actions. Note, these mappings do not have to be run in tandem: $G_i$ can be run at a different frequency than $S$ if needed. In practice, $G_i$ almost always runs at a higher frequency to perform low-level control.

We implement $S$ using a modified VLA. Given the current observation, $o_t$, the model predicts the VLA target $g_{t_\text{VLA}} \in G_\text{VLA}$, the skill index, $i_t$, and the target information for each classical skill $\{g_{t_0}, \; g_{t_1}, \dots, \; g_{t_n}\}$.
\begin{equation}
    i, \; g_{t_\text{VLA}}, \; g_{t_0}, \; g_{t_1}, \dots, \;  g_{t_n} = \pi_\theta(o_t)
\end{equation}

The selected controller $C_i$ is then used to calculate the action to send the robot, $a_t = C_{i}(o_t, g_{t_i})$,
where the skill index $i$, can point to either one of the classical skills or the VLA action head. In fact, under this formulation, the VLA action head is treated identically to the classical skills: the model predicts some target information (in this case, an action chunk of delta goal poses), which is then sent to a low-level controller to be executed on the robot. Therefore, we view our proposed system not as a post-hoc addition to a VLA, but rather a natural extension of the VLA's action space. The action chunk returned by a standard VLA is converted to motor torques by a low-level controller, usually an impedance-based reference tracker. Our system simply provides alternative low-controllers for the VLA to choose from, and modifies the model's action space accordingly; there is nothing unique about the default VLA controller. However, for practical reasons, our implementation does treat the VLA action head differently from the rest of the target head outputs. Specifically, this action head is included in the pretrained VLA and predicts a trajectory rather than a single target, so it is treated independently from the classical skill target head.

\section{Method}

\subsection{System Overview}

\begin{figure}[t!]
\centering
\includegraphics[width=\linewidth]{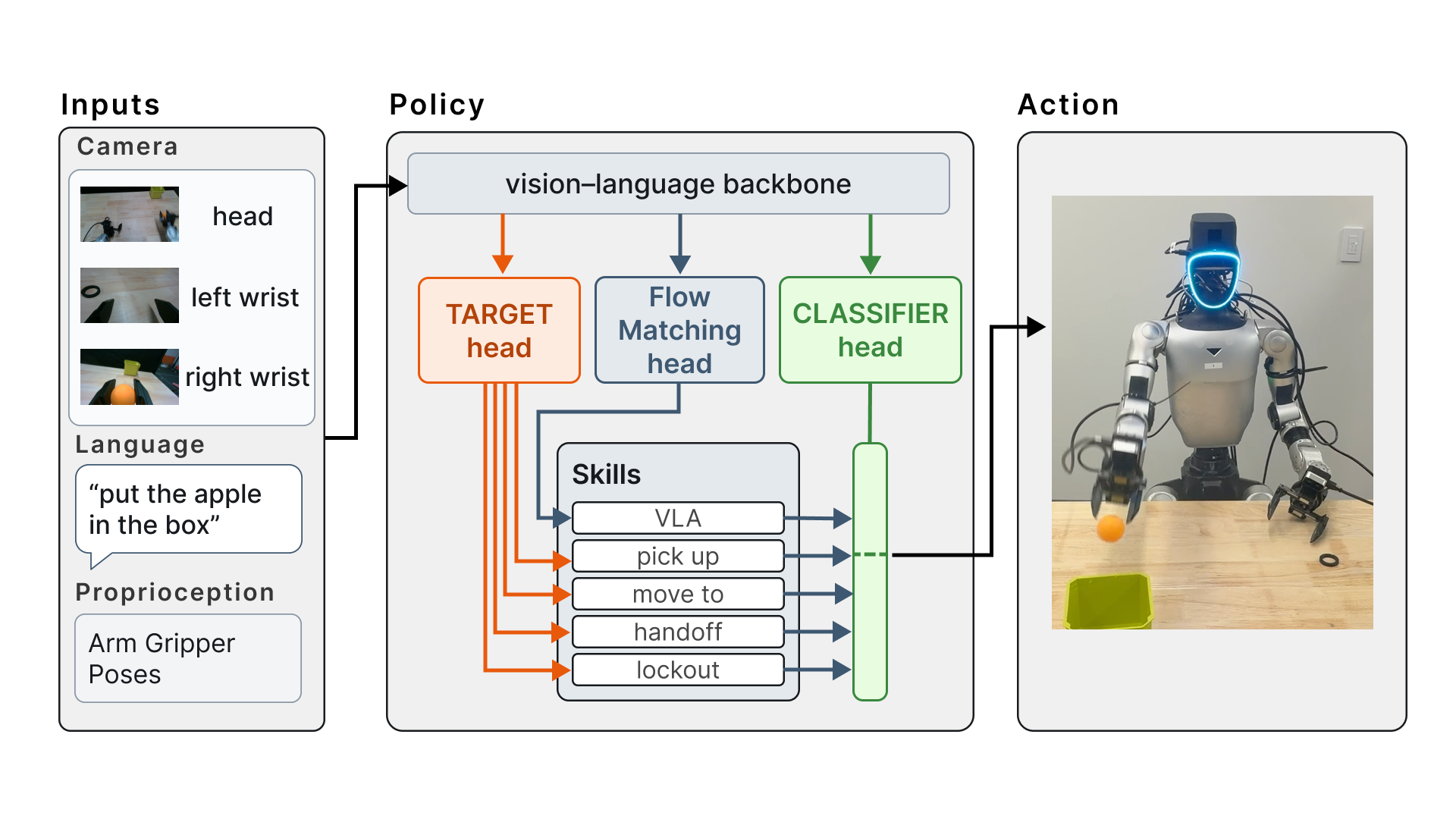}
\caption{SkilledVLA system overview. The VLA backbone consumes camera observations, language instruction, and proprioception; its pooled
features feed the flow-matching action head and the auxiliary heads. The
\textit{classifier} head determines which controller to use, and the \textit{target}
heads output conditioning information for each controller.}
\label{fig:system}
\end{figure}
SkilledVLA modifies a pretrained VLA to expose two additional outputs (Fig.~\ref{fig:system}): a \textit{classifier} that determines which skill (either a classical skill or the VLA) should be used to control each arm, and series of \textit{targets}, which provide task-relevant information such as which object a skill should interact with, the goal position the robot should move to, or a retreat position for the skill to terminate at (e.g., in handoff, \textit{target} tells the hands where to go after the exchange so the VLA can continue from an in-distribution location). SkilledVLA runs without kinematics in the policy stack: the policy emits Cartesian wrist poses that the robot's existing Inverse Kinematic (IK) retargeter consumes.

\subsection{Classical Skills}

Table~\ref{tab:skills} lists the skill vocabulary. The skills are as follows: \skill{VLA} keeps an arm under GR00T control, \skill{lockout} holds an arm in place, locking all joints, \skill{pick\_up} uses a 3D point target to select which object to grasp (the object, segmented by SAM, closest to the point is selected), \skill{move\_to} in simulation plans a collision-free path with cuRobo~\cite{sundaralingam2023curobo}, \skill{move\_to} on the real robot streams a rate-limited interpolation of the wrist pose toward the \textit{target} pose, \skill{insertion} servos a held part into a hole and pushes the object into the found hole, \skill{handoff} is a bimanual skill that has one gripper grasp an object out of the other gripper, and \skill{toss} is a ballistic task that throws an object to a target location. SkilledVLA was deployed both in simulation and on a real robot. The \skill{insertion} skill was only used in simulation, and \skill{toss} was only used on the real robot. All other skills were available on all platforms.

For \skill{pick\_up} on the real robot, we segment the object using SAM conditioned on the \textit{target} point. We then enumerate and score antipodal contact pairs on the segmented point cloud, and execute the decided grasp. For simulated dexterous end-effectors, where this parallel-jaw formulation does not apply, we use manually tuned, object-specific grasps. These hand-designed grasps are deliberately a simple stand-in for a general dexterous grasp-synthesis module: SkilledVLA's dispatch interface is agnostic to what fills the grasp slot, and a stronger planner would easily drop in. \skill{insertion} captures the hole pose and the tool-in-hand transform once after grasp, aligns the part's axis to the hole axis within a lateral and angular budget, then inserts. A slip monitor re-captures the grasp transform if it drifts by more than 6 mm, jams trigger a back-off and retry, and a spiral search recovers centering in the final millimeters of insertion if necessary. To insert into open-top receptacles, the skill recenters above the mouth, descends, and releases when concentric to within 5 mm, letting the part self-seat. \skill{handoff} brings the arms to a meet pose, closes the receiving gripper on the object from the giving gripper, verifies that the object is grasped in the receiving gripper, releases the giving gripper, and finally parks both arms at their \textit{target} end poses. These end poses are necessary because the meet pose itself resembles the trained trigger state, and terminating immediately after handoff can cause the arms to cyclically pass the object back and forth between the two hands.
\skill{toss} segments the object rim specified by \textit{target} and then solves closed-form ballistics to determine the release pose and velocity required to launch a projectile to the center of the rim, and finally conducts the smooth motion toss.
\subsection{Auxiliary Heads}

The \textit{classifier} and \textit{target} heads are two-layer MLPs that read the VLA backbone's pooled vision--language and state features. The \textit{classifier} heads (one for each arm) are trained to predict which skill (including the base VLA) should control each arm, using annotated per-frame class labels. At deployment, a small debounce filter is applied to limit erroneous mode switches, requiring two consecutive inference calls to change controller. The \textit{target} head predicts a 9-D target (position and two rotation matrix rows~\cite{zhou2019continuity}) under a masked L1 loss that is active only inside goal-bearing skill windows. The prediction is used in three ways depending on the skill: for \skill{pick\_up} and \skill{toss} the target's position is used to provide a 3-D point locating the target; for \skill{handoff} the target specifies the full 9-D pose the arm should occupy after the exchange; and for \skill{move\_to} the target is the 9-D goal pose of the end-effector.
\begin{equation}
    \mathcal{L} = \mathcal{L}_{\mathrm{FM}} + \,\mathrm{CE}\!\left(h_{\mathrm{\textit{classifier}}}(\mathrm{sg}(z))\right) + \lambda\,\mathrm{L1}\!\left(h_{\mathrm{\textit{target}}}(\mathrm{sg}(z))\right)
\end{equation}
where $\lambda = 10$ is a weighting term, $z$ is the pooled backbone feature and $\mathrm{sg}$ is the stop-gradient operator: the backbone is fine-tuned by the default flow-matching action loss alone - the auxiliary losses do not backpropagate into the VLM backbone. For training, ground truth labels for skill classification were annotated on the demonstration trajectory. This annotation was done by a human operator, with assistance from a VLM (Claude Opus 4.8). Although a VLM was used to accelerate the labeling process, all annotations were reviewed by a human.

\begin{table}[t]
\renewcommand{\arraystretch}{1.15}
\caption{Skill taxonomy. Goal-bearing classes are the ones inside whose
labeled windows the \textit{target} head is supervised.}
\label{tab:skills}
\centering
\footnotesize
\setlength{\tabcolsep}{3.5pt}
\begin{tabular}{@{}llccc@{}}
\hline
Class & Controller & Target & Sim & Real \\
\hline
\skill{none}      & policy (GR00T)              & \cmark           & \cmark & \cmark \\
\skill{pick\_up}  & grasp then lift             & \cmark   & \cmark & \cmark \\
\skill{move\_to}  & collision-free reach        & \cmark   & \cmark & \cmark \\
\skill{insertion} & align gate, spiral search   & \cmark   & \cmark & \xmark \\
\skill{handoff}   & force-verified exchange     & \cmark & \cmark & \cmark \\
\skill{lockout}   & hold pose                   &  \xmark            & \cmark & \cmark \\
\skill{toss}      & ballistic release           & \cmark   &  \xmark          & \cmark \\
\hline
\end{tabular}
\end{table}

\subsection{Dispatch, Arbitration, and Handback}

At each inference call, the \textit{classifier} head predicts which skill to execute for each arm. A skill is selected if it wins the per-arm argmax for two consecutive inference calls; without this debounce, single-chunk mispredictions could dispatch spurious skills. For safety and to ensure any assumptions made by the classical skills are honored, each skill has a set of pre-conditions that are checked before the skill is called. Object-centric vision-based skills such as \skill{pick\_up} verify that the desired object exists in the scene (minimum segmentation score), hand-off and throw skills verify that an object is in the gripper, and movement skills clamp the goal position to lie within the workspace. Additionally, the \skill{toss} skill requires the target opening to be larger than the thrown object and lie within an achievable ballistic envelope. Bimanual skills have an additional rule: both arms' \textit{classifier} must select the same bimanual skill.

In addition to pre-conditions, the skills have termination conditions that apply safeguards to the termination of the skill. The first of these conditions is a stuck-abort clause: if end effectors get stuck in a location, or the skill exceeds a maximum duration, we abort the skill and instead pick the next highest-scoring skill from the \textit{classifier}, with the debounce filter ensuring that the skill is not resumed for at least one inference step.

\begin{figure*}[t!]
\centering
\includegraphics[width=\textwidth]{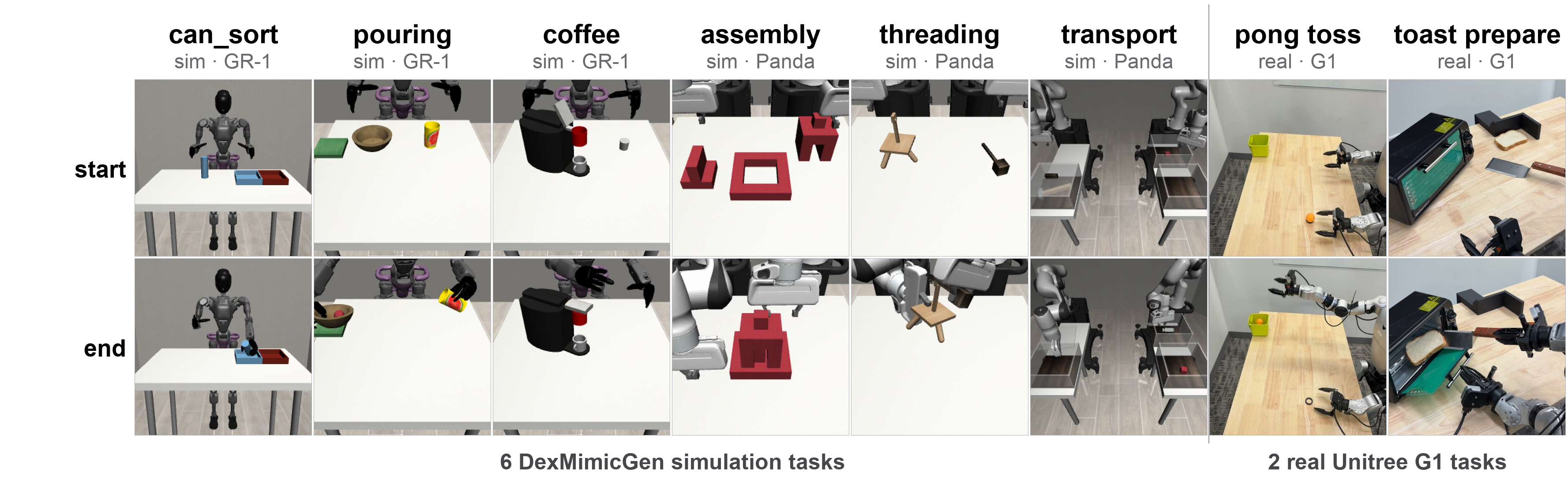}
\caption{The eight evaluation tasks, showing the scene at the first
step (top row) and the last step (bottom row) of each task. The left three columns are GR-1 based tasks, and the following three columns are bimanual Panda tasks from DexMimicGen
simulation tasks. The right two images are real Unitree G1 tasks on a Unitree G1 with Dex1 grippers.
}
\label{fig:tasks}
\end{figure*}

\section{Experimental Setup}

We evaluated this system on six simulation tasks and two real Unitree G1 tasks. The six simulation tasks were taken directly from DexMimicGen\cite{jiang2025dexmimicgen}:
 transport, can\_sort, assembly, coffee, pouring, and threading. The two real tasks are \emph{pong toss}, in which a ball must be picked up and thrown in a box, and \emph{toast prepare}, in which the robot must pick up a piece of bread using a spatula, open a toaster oven, place the bread into the toaster oven, and close the oven door. \emph{Pong toss} was chosen because it is a high-precision dynamic task, and \emph{toast prepare} was chosen because it is a long horizon task that requires tool manipulation and steady end-effector movements.
Fig.~\ref{fig:tasks} shows the initial and final state of one episode of each task.

\subsection{Real-Robot Setup}
\label{sec:real-setup}
We conducted hardware experiments on a Unitree G1 humanoid with Dex1 grippers, with a head mounted Intel RealSense D435 camera, and two Intel RealSense D405 cameras mounted on the top of each wrist. The policy reads a 23-D state (9-D pose per wrist, three waist angles, and two gripper positions) and emits 20-D end-effector actions (a 9-D wrist pose and gripper command per arm) in 16-step chunks at 20~Hz. End-effector actions are predicted as relative goal poses, measured in the end-effector's current pose, and gripper actions are absolute width; the policy executor re-anchors each chunk against the recorded robot state, so commands reach the robot as absolute end-effector poses.
These goal poses are sent to the robot's built-in controller, which tracks the references using a gravity-compensated impedance controller. Hardware experiments use the skills marked in the Real column of Table~\ref{tab:skills}.
Skills that require object information (such as \skill{pick\_up}) use RGB-D images from the head camera, with SAM \cite{sam3} for object segmentation (conditioned on the \textit{target} point). We evaluate on two bimanual tasks. In \emph{pong toss}, a ball is picked up with the left hand, handed to the right, and thrown into a box; we score pick up, handoff, and toss, the last representing task success. In \emph{toast prepare}, the right end-effector picks up the spatula and scoops a slice of bread off of the table.
Then, the left end effector opens up a toaster oven by grasping its handle, and the right end effector uses the spatula to place the bread in the toaster, then removes the spatula (leaving the bread in the oven). Finally, the robot closes the toaster oven door with the left end effector.
The task is marked as successful if the robot places the toast into the toaster using the spatula, and closes the oven door. For each task, a teleoperator collected 50 successful demonstrations, with 45 of these demonstrations being used for model fine-tuning and 5 withheld for offline validation. During experimental evaluation, each task was run 20 times. For the pong task, the start locations of the ping pong ball and box were randomized within a 15 cm $\times$ 15 cm area, and for the \emph{toast prepare} the spatula was randomly placed within an 8 cm $\times$ 4 cm area, with a rotation randomization of $\pm20^{\circ}$. The size of the randomization area was chosen to ensure that the task could be completed within the robot's workspace.
\subsection{Simulation Setup}
\label{sec:sim-setup}

The simulation setup consisted of six DexMimicGen~\cite{jiang2025dexmimicgen} bimanual tasks across two embodiments, simulated in MuJoCo~\cite{todorov2012mujoco} via robosuite~\cite{zhu2020robosuite}. \emph{Threading}, \emph{Three-Piece Assembly}, and \emph{Transport} use two Panda arms with parallel plate end-effectors, controlled using per-arm OSC pose deltas plus gripper width (14-D), with a 20-D observation space. \emph{Coffee}, \emph{Pouring}, and \emph{Can Sort} run on the GR-1 humanoid with dexterous hands, controlled with a 24-D action space of absolute pose targets and six finger commands per hand, and a 30-D observation space.

In \emph{Threading}, the needle's bar must pass through the tripod's ring, with the needle and tripod starting on opposite sides of the table and neither fixed down. In \emph{Three-Piece Assembly}, the robot must first seat piece\_1 into the base, then place piece\_2 on top of piece\_1 and release all objects from contact with the grippers; the base spawns in the center of the table with the two pieces on opposite sides of it, one piece in each arm's reach. In \emph{Transport}, the robot must move the hammer from the start bin to the target bin and move the trash from the target bin to the trash bin; the start bin is covered by a lid that the robot must remove first, and the two bins sit on opposite sides so the hammer must change hands. In \emph{Coffee}, the robot must pick up a coffee pod, place it in the top of a coffee machine, and then shut the machine's lid over the pod. In \emph{Pouring}, the robot must pick up and pour a ball out of a cup and into a bowl, and place the bowl (with ball inside) onto a green pad. Finally, in \emph{Can Sort}, the robot must place the can into the bin matching the can's color, which is randomized every episode; the bins are reachable only by the left arm and the can always spawns within the right arm's reach, so it has to change hands on the way across.

For each task, 300 generated demonstrations (supplied by DexMimicGen) were used to finetune the VLAs. Evaluation follows the benchmark's protocol: 50 rollouts per task using randomized object placements and success verification done via the environment's success check, with a 1000-step cap. When comparing controls, we use the same initial seed across the different methods, so that all evaluations are completed on the same set of randomized initial conditions.
For simplicity, we allow classical skills to use the simulator state for object detection, as this perception challenge is separate from our primary investigation, which is into the integration of classical skills into VLAs.

\subsection{Baselines}

We compare against three similar architectures, both run under the protocol described above. The first, vanilla GR00T, uses the same pretrained GR00T checkpoint used in SkilledVLA without the additional skill-related heads. We use the same training data with identical hyperparameters for Vanilla GR00T and SkilledVLA. The second method,
$\pi_{0.5}$-H is a re-implementation of the two-level model~\cite{intelligence2025pi_}, built on its authors' openpi codebase~\cite{openpi2024}. Its high-level supervision is the same per-frame skill labels that supervise our \textit{classifier} head, rendered as short language phrases, so both models receive identical supervision. At evaluation it is self-contained: it decodes its own subtask, with no GR00T and no external \textit{classifier} or \textit{target} head in the loop. We fine-tune from the $\pi_{0.5}$ base checkpoint per task and match every axis of the comparison to the GR00T columns: eight-step replan cadence, deterministic sampling. We also raise its batch size so it sees 640k samples, the same as GR00T.

The VLM orchestrator implementation uses GPT-6 Astra in a Pigey-style~\cite{galanti2026addressing} tool-calling loop using the skill library from SkilledVLA. In addition to these skills, we provide the VLM with utility skills (a raw-coordinate move, release, and retreat), a VLA rollout tool, which runs the fine-tuned vanilla GR00T for a fixed horizon, and a termination skill (``done''). For object based skills, such as \skill{pick\_up}, we replace the point conditioning with text conditioning for SAM - the VLM simply states the name of the object it wishes to interact with, as this is a natural interaction mechanism for the VLM. In simulation, the VLM is grounded using oracle information of each task's scene from the simulator. It is prompted as the high-level controller of a two-armed robot and emits one tool call per invocation. Before every call it receives the three camera views and a state header carrying every object's world pose and each arm's grasp flag, and returns the skill it wishes to use and the required inputs for the skill. Every call returns a typed result (success or a named failure mode), so the orchestrator replans after every skill rather than on a timer.

\section{Results}
We compare SkilledVLA against vanilla fine-tuned GR00T, a hierarchical VLA ($\pi_{0.5}$), and a VLM orchestrator on two real Unitree G1 tasks (Table~\ref{tab:real}) and six simulated tasks across two embodiments (Table~\ref{tab:sim}). We then analyze where failures occur.
\begin{table}[t]
\renewcommand{\arraystretch}{1.15}
\caption{Simulation comparison: successes out of 50 rollouts per task}
\label{tab:sim}
\centering
\footnotesize
\begin{tabular}{@{}llcccc@{}}
\hline
Task & Embod. & Vanilla & SkilledVLA & $\pi_{0.5}$-H & VLM-orch \\
\hline
can\_sort  & GR-1  & \textbf{100\%} & \textbf{100\%} & 88\% & 98\% \\
assembly   & Panda & 62\% & \textbf{80\%} & 14\% & 38\%  \\
threading  & Panda & 42\% & 82\% & 10\% & \textbf{84\%} \\
transport  & Panda & 26\% & \textbf{36\%} & 32\% & 8\%  \\
pouring    & GR-1  & 70\% & \textbf{94\%} & 74\% & 0\%  \\
coffee     & GR-1  & 20\% & \textbf{54\%} & 48\% & 6\%  \\
\hline
total      &       & 53\% & \textbf{74\%} & 44\% & 39\% \\
\hline
\end{tabular}
\end{table}

\subsection{Real-Robot Comparison}

\begin{figure*}[t!]
\centering
\includegraphics[width=\textwidth]{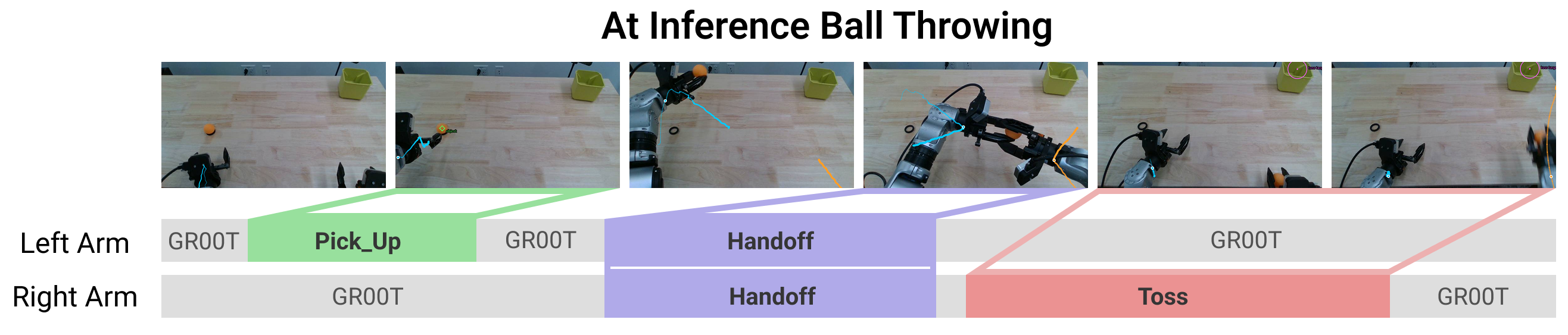}
\caption{One SkilledVLA rollout of the real pong-toss task (head-camera frames,
left to right). The bands below each lane show which controller drives each
arm at every moment: GR00T by default (white), then the \textit{classifier} head dispatches
\skill{pick\_up} on the left arm (green), a bimanual \skill{handoff} that
fires on both arms over the same window (purple), and a ballistic
\skill{toss} on the right arm (red), after which control returns to GR00T.
End-to-end success on this task rises from 5\% to 60\% (1/20 to 12/20).}
\label{fig:pong}
\end{figure*}

\begin{table}[t]
\renewcommand{\arraystretch}{1.15}
\caption{Real-robot comparison on the Unitree G1: overall success rate on 20 trials per task and stage-wise performance}
\label{tab:real}
\centering
\footnotesize
\setlength{\tabcolsep}{4.5pt}
\begin{tabular}{@{}lcccc@{}}
\hline
Stage & Vanilla & SkilledVLA & $\pi_{0.5}$-H & VLM-orch \\
\hline
\multicolumn{5}{@{}l}{\emph{pong toss}} \\
\quad pick up & 20/20 & 16/20 & 19/20 & 20/20 \\
\quad handoff & 17/20 & 16/16 & 4/19  & 17/20 \\
\quad toss    & 1/17  & 12/16 & 0/4   & 10/17 \\
\hline
\quad \textbf{overall}   & 5\%   & \textbf{60\%} & 0\% & 50\% \\
\hline
\hline
\multicolumn{5}{@{}l}{\emph{toast prepare}} \\
\quad pick up & 19/20 & 19/20 & 13/20 & 14/20 \\
\quad open  & 12/19 & 17/19 & 5/13  & 10/14 \\
\quad load  & 5/12  & 10/17 & 1/5   & 2/10  \\
\hline
\quad \textbf{overall} & 20\% & \textbf{50\%} & 5\% & 10\% \\
\hline
\end{tabular}
\end{table}
SkilledVLA improves upon vanilla-finetuned GR00T, a hierarchical VLA ($\pi_{0.5}$), and a VLM orchestrator on both real tasks (Fig.~\ref{fig:pong} shows one rollout of SkilledVLA). SkilledVLA succeeds in 60\% and 50\% of trials compared to vanilla GR00T's 5\% and 20\%, $\pi_{0.5}$'s 0\% and 5\%, and VLM orchestrator's 50\% and 10\% on the \emph{pong toss} and \emph{toast prepare} tasks respectively.

Both real tasks are multi-stage and bimanual: \emph{pong toss} chains a pick up, a cross-arm handoff, and a ballistic toss, and \emph{toast prepare} chains tool use, oven-door manipulation, transport of the slice, placement, and closing the door. Throughout, we report each stage's success out of the trials that actually reached that stage, since an upstream failure means a downstream stage is never attempted. On \emph{pong toss}, vanilla GR00T completes every pick up and most handoffs, but converts only 1 of the 17 completed handoffs into a successful toss.
This failure is expected, as tossing a ball into a container requires well-timed, coordinated position and velocity control, which is inherently difficult for VLAs to achieve, due to their relatively low re-planning frequency and goal-pose action chunk representation. SkilledVLA, by contrast, was able to directly solve the ballistics problem and use a high-frequency control loop to achieve the desired launch position and velocity.

$\pi_{0.5}$ also had a very high success rate for the pick up task, but performed very poorly at the handoff stage and even worse at the toss stage with 0 successes. During handoff there was no semblance of grasp verification and as such the giving gripper tended to open before the receiving gripper was able to close. VLM orchestrator's pick up phase was also successful every time, but struggled more than SkilledVLA at the handoff stage due to poor error recovery. In addition to dropping the ball during this stage, poorer handoff performance leads to lower quality grasps, resulting in a lower toss success rate. SkilledVLA's own pong failures are at pick up, which failed in 20\% of the time, likely due to a \textit{target} head that was not precise at pointing directly onto the pong ball; of the  trials that completed the pick up though, all also completed handoff.

On the \emph{toast prepare} task, vanilla GR00T's failures mostly lie in its ability to grasp the handle of the toaster oven while keeping the bread on the spatula. This is largely attributable to the fact that the VLA tends to move the spatula hand while grasping the handle which drops the toast. In SkilledVLA, grasping the handle is its own specific skill call to the pick\_up skill, during which the spatula arm is locked out, preventing any shaking while the oven door is being opened. VLM orchestrator's low success rate in depositing the toast in the toaster oven lies in it terminating the GR00T runner skill too soon. GR00T is often slow to dump the bread, and the VLM struggled to predict how long to run the VLA. SkilledVLA, by contrast, is much better at determining when to switch skills. $\pi_{0.5}$'s main struggle was in its ability to grasp objects, whether it be the spatula or the toaster oven's handle. Each of these grasps require very specific styles of grasps to ensure future success, which $\pi_{0.5}$ struggled to do. Subsequent failures for SkilledVLA lay in its ability to drop the toast in the oven. Even with a proper grasp of the oven handle, dropping the toast in properly is a mix of navigating to the oven without hitting the spatula and toast on anything along the way and properly tilting to release the toast.

\subsection{Simulation Comparison and Ablations}

Across the six tasks (Table~\ref{tab:sim}), SkilledVLA wins or ties every task when compared against vanilla GR00T, $\pi_{0.5}$-H, and the VLM orchestrator. Against vanilla GR00T, SkilledVLA's largest gains happen in the tasks with very small tolerances, such as the threading and coffee tasks. The tasks where vanilla GR00T came close to SkilledVLA's performance were tasks that did not require as much fine-grained manipulation and had looser tolerances, such as the can sort and assembly tasks.
$\pi_{0.5}$-H was close to the success rate of SkilledVLA on the GR-1 tasks, and trailed behind on the Panda arm tasks. Both models received identical supervision and the decoded subtasks match the labels on the held-out frames, so the failure cannot lie in the prediction of the phases. Additionally, the gap concentrates in $\pi_{0.5}$-H's low-level execution on the Panda embodiment, especially on tasks with fine-grained insertions, suggesting that the learned low-level skills under-perform their classical counterparts.

The VLM as orchestrator had access to vanilla GR00T and the same skills as SkilledVLA, yet on most tasks scored worse than both. The VLM lacks the ability to compose a proper flow of skill calling to complete tasks. The skills mostly succeed on their own when they are called at the right time.
The orchestrator does especially poorly on pouring and coffee, succeeding in 0\% and 6\% of rollouts, which is attributable to the orchestrator failing to approach the already fragile grasp positions for the coffee pod and cup. In SkilledVLA this is not a failure because the pick up skills are only called after the VLA navigates the end effectors to certain approach poses. The single task that VLM as orchestrator beats all other methods is on the threading task, which relies heavily on the threading skill's robustness. Having access to such a robust skill gives VLM as orchestrator the best success rate.

\section{Conclusion and Limitations}

We presented SkilledVLA, a system to give VLAs access to classical skills. By adding a \textit{classifier} head to select which skill to invoke and a \textit{target} head to condition the skills, our method directly integrates skills into a VLA's action space, combining the generalizability of large scale end-to-end learning with the reliability and performance of engineered skills. We validated SkilledVLA against three other similar methods, a vanilla finetuned VLA (GR00T), a hierarchical VLA ($\pi_{0.5}$), and a VLM orchestrator. On a real humanoid robot and in simulation SkilledVLA has a higher success rate than all other methods.

SkilledVLA's benefit lies in its ability to call on classical skills at a VLA's inference speed; though each skill must be individually engineered. In many cases these classical skills already exist for systems, but in the case they do not, time will need to be spent to make them.
At the training stage these skills must also be predefined; therefore, a skill cannot be added on after the fine-tuning process unless the VLA and auxiliary heads are re-fine-tuned. The auto-labeling of episodes is not a very time-intensive task due to the capabilities of modern LLMs, but it does still require some additional effort. Dexterous-hand grasping also still remains a challenging realm, though the challenges with this can be addressed in the future by more well designed dexterous manipulation grasping systems and techniques that this paper did not focus on.

In the future, we would be interested to see how increasing skill library size impacts the system's successes. Currently, SkilledVLA is limited to manipulation, but  investigation into how such methods may carry over to locomotion could prove fruitful. Additionally, using reinforcement learning to further fine-tune such a system's additional skill related heads, would be an interesting area for exploration.

\bibliographystyle{IEEEtran}
\bibliography{references}

\begin{thebibliography}{10}
\providecommand{\url}[1]{#1}
\csname url@samestyle\endcsname
\providecommand{\newblock}{\relax}
\providecommand{\bibinfo}[2]{#2}
\providecommand{\BIBentrySTDinterwordspacing}{\spaceskip=0pt\relax}
\providecommand{\BIBentryALTinterwordstretchfactor}{4}
\providecommand{\BIBentryALTinterwordspacing}{\spaceskip=\fontdimen2\font plus
\BIBentryALTinterwordstretchfactor\fontdimen3\font minus \fontdimen4\font\relax}
\providecommand{\BIBforeignlanguage}[2]{{%
\expandafter\ifx\csname l@#1\endcsname\relax
\typeout{** WARNING: IEEEtran.bst: No hyphenation pattern has been}%
\typeout{** loaded for the language `#1'. Using the pattern for}%
\typeout{** the default language instead.}%
\else
\language=\csname l@#1\endcsname
\fi
#2}}
\providecommand{\BIBdecl}{\relax}
\BIBdecl

\bibitem{ahn2022can}
M.~Ahn, A.~Brohan, N.~Brown, Y.~Chebotar, O.~Cortes, B.~David, C.~Finn, C.~Fu, K.~Gopalakrishnan, K.~Hausman \emph{et~al.}, ``Do as I can, not as I say: Grounding language in robotic affordances,'' \emph{arXiv preprint arXiv:2204.01691}, 2022.

\bibitem{liang2023code}
J.~Liang, W.~Huang, F.~Xia, P.~Xu, K.~Hausman, B.~Ichter, P.~Florence, and A.~Zeng, ``Code as policies: Language model programs for embodied control,'' in \emph{2023 IEEE International Conference on Robotics and Automation (ICRA)}.\hskip 1em plus 0.5em minus 0.4em\relax IEEE, 2023, pp. 9493--9500.

\bibitem{huang2023voxposer}
W.~Huang, C.~Wang, R.~Zhang, Y.~Li, J.~Wu, and L.~Fei-Fei, ``VoxPoser: Composable 3D value maps for robotic manipulation with language models,'' \emph{arXiv preprint arXiv:2307.05973}, 2023.

\bibitem{intelligence2025pi_}
{Physical Intelligence}, K.~Black, N.~Brown, J.~Darpinian, K.~Dhabalia, D.~Driess, A.~Esmail, M.~Equi, C.~Finn, N.~Fusai \emph{et~al.}, ``$\pi_{0.5}$: A vision-language-action model with open-world generalization,'' \emph{arXiv preprint arXiv:2504.16054}, 2025.

\bibitem{belkhale2023hydra}
S.~Belkhale, Y.~Cui, and D.~Sadigh, ``HYDRA: Hybrid robot actions for imitation learning,'' in \emph{Conference on Robot Learning}.\hskip 1em plus 0.5em minus 0.4em\relax PMLR, 2023, pp. 2113--2133.

\bibitem{sundaresan2025s}
P.~Sundaresan, H.~Hu, Q.~Vuong, J.~Bohg, and D.~Sadigh, ``What's the move? Hybrid imitation learning via salient points,'' in \emph{International Conference on Learning Representations}, 2025.

\bibitem{bjorck2025gr00t}
J.~Bjorck, F.~Casta{\~n}eda, N.~Cherniadev, X.~Da, R.~Ding, L.~Fan, Y.~Fang, D.~Fox, F.~Hu, S.~Huang \emph{et~al.}, ``GR00T N1: An open foundation model for generalist humanoid robots,'' \emph{arXiv preprint arXiv:2503.14734}, 2025.

\bibitem{black2024pi_0}
K.~Black, N.~Brown, D.~Driess, A.~Esmail, M.~Equi, C.~Finn, N.~Fusai, L.~Groom, K.~Hausman, B.~Ichter \emph{et~al.}, ``$\pi_0$: A vision-language-action flow model for general robot control,'' \emph{arXiv preprint arXiv:2410.24164}, 2024.

\bibitem{kim2024openvla}
M.~J. Kim, K.~Pertsch, S.~Karamcheti, T.~Xiao, A.~Balakrishna, S.~Nair, R.~Rafailov, E.~Foster, G.~Lam, P.~Sanketi \emph{et~al.}, ``OpenVLA: An open-source vision-language-action model,'' \emph{arXiv preprint arXiv:2406.09246}, 2024.

\bibitem{chi2025diffusion}
C.~Chi, Z.~Xu, S.~Feng, E.~Cousineau, Y.~Du, B.~Burchfiel, R.~Tedrake, and S.~Song, ``Diffusion policy: Visuomotor policy learning via action diffusion,'' \emph{The International Journal of Robotics Research}, vol.~44, no. 10-11, pp. 1684--1704, 2025.

\bibitem{lipman2022flow}
Y.~Lipman, R.~T.~Q. Chen, H.~Ben-Hamu, M.~Nickel, and M.~Le, ``Flow matching for generative modeling,'' \emph{arXiv preprint arXiv:2210.02747}, 2022.

\bibitem{belkhale2024rt}
S.~Belkhale, T.~Ding, T.~Xiao, P.~Sermanet, Q.~Vuong, J.~Tompson, Y.~Chebotar, D.~Dwibedi, and D.~Sadigh, ``RT-H: Action hierarchies using language,'' \emph{arXiv preprint arXiv:2403.01823}, 2024.

\bibitem{shi2025hi}
L.~X. Shi, B.~Ichter, M.~Equi, L.~Ke, K.~Pertsch, Q.~Vuong, J.~Tanner, A.~Walling, H.~Wang, N.~Fusai \emph{et~al.}, ``Hi Robot: Open-ended instruction following with hierarchical vision-language-action models,'' \emph{arXiv preprint arXiv:2502.19417}, 2025.

\bibitem{zhai2026skillvla}
X.~Zhai, Z.~Huang, L.~Wu, Q.~Zhao, Q.~Yu, J.~Ren, C.~Hao, and H.~Soh, ``SkillVLA: Tackling combinatorial diversity in dual-arm manipulation via skill reuse,'' \emph{arXiv preprint arXiv:2603.03836}, 2026.

\bibitem{hu2023look}
Y.~Hu, F.~Lin, T.~Zhang, L.~Yi, and Y.~Gao, ``Look before you leap: Unveiling the power of GPT-4V in robotic vision-language planning,'' \emph{arXiv preprint arXiv:2311.17842}, 2023.

\bibitem{galanti2026addressing}
L.~Galanti, D.~Shah, and T.~Dao, ``Addressing the orchestration gap in generalist robots via physical agency,'' \emph{arXiv preprint arXiv:2607.21725}, 2026.

\bibitem{yao2022react}
S.~Yao, J.~Zhao, D.~Yu, N.~Du, I.~Shafran, K.~Narasimhan, and Y.~Cao, ``ReAct: Synergizing reasoning and acting in language models,'' \emph{arXiv preprint arXiv:2210.03629}, 2022.

\bibitem{sundaralingam2023curobo}
B.~Sundaralingam, S.~K.~S. Hari, A.~Fishman, C.~Garrett, K.~Van~Wyk, V.~Blukis, A.~Millane, H.~Oleynikova, A.~Handa, F.~Ramos \emph{et~al.}, ``cuRobo: Parallelized collision-free robot motion generation,'' in \emph{2023 IEEE International Conference on Robotics and Automation (ICRA)}.\hskip 1em plus 0.5em minus 0.4em\relax IEEE, 2023, pp. 8112--8119.

\bibitem{zeng2020tossingbot}
A.~Zeng, S.~Song, J.~Lee, A.~Rodriguez, and T.~Funkhouser, ``TossingBot: Learning to throw arbitrary objects with residual physics,'' \emph{IEEE Transactions on Robotics}, vol.~36, no.~4, pp. 1307--1319, 2020.

\bibitem{nguyen1988constructing}
V.-D. Nguyen, ``Constructing force-closure grasps,'' \emph{The International Journal of Robotics Research}, vol.~7, no.~3, pp. 3--16, 1988.

\bibitem{ferrari1992planning}
C.~Ferrari and J.~Canny, ``Planning optimal grasps,'' in \emph{Proceedings 1992 IEEE International Conference on Robotics and Automation}, vol.~3.\hskip 1em plus 0.5em minus 0.4em\relax IEEE, 1992, pp. 2290--2295.

\bibitem{ten2017grasp}
A.~Ten~Pas, M.~Gualtieri, K.~Saenko, and R.~Platt, ``Grasp pose detection in point clouds,'' \emph{The International Journal of Robotics Research}, vol.~36, no. 13-14, pp. 1455--1473, 2017.

\bibitem{chhatpar2001search}
S.~R. Chhatpar and M.~S. Branicky, ``Search strategies for peg-in-hole assemblies with position uncertainty,'' in \emph{Proceedings 2001 IEEE/RSJ International Conference on Intelligent Robots and Systems. Expanding the Societal Role of Robotics in the the Next Millennium (Cat. No. 01CH37180)}, vol.~3.\hskip 1em plus 0.5em minus 0.4em\relax IEEE, 2001, pp. 1465--1470.

\bibitem{zhou2019continuity}
Y.~Zhou, C.~Barnes, J.~Lu, J.~Yang, and H.~Li, ``On the continuity of rotation representations in neural networks,'' in \emph{2019 IEEE/CVF Conference on Computer Vision and Pattern Recognition (CVPR)}.\hskip 1em plus 0.5em minus 0.4em\relax IEEE, 2019, pp. 5738--5746.

\bibitem{jiang2025dexmimicgen}
Z.~Jiang, Y.~Xie, K.~Lin, Z.~Xu, W.~Wan, A.~Mandlekar, L.~J. Fan, and Y.~Zhu, ``DexMimicGen: Automated data generation for bimanual dexterous manipulation via imitation learning,'' in \emph{2025 IEEE International Conference on Robotics and Automation (ICRA)}.\hskip 1em plus 0.5em minus 0.4em\relax IEEE, 2025, pp. 16\,923--16\,930.

\bibitem{sam3}
N.~Carion, L.~Gustafson, Y.-T. Hu, S.~Debnath, R.~Hu, D.~Suris, C.~Ryali, K.~V. Alwala, H.~Khedr, A.~Huang \emph{et~al.}, ``{SAM 3}: Segment anything with concepts,'' \emph{arXiv preprint arXiv:2511.16719}, 2025.

\bibitem{todorov2012mujoco}
E.~Todorov, T.~Erez, and Y.~Tassa, ``MuJoCo: A physics engine for model-based control,'' in \emph{2012 IEEE/RSJ International Conference on Intelligent Robots and Systems}.\hskip 1em plus 0.5em minus 0.4em\relax IEEE, 2012, pp. 5026--5033.

\bibitem{zhu2020robosuite}
Y.~Zhu, J.~Wong, A.~Mandlekar, R.~Mart{\'\i}n-Mart{\'\i}n, A.~Joshi, K.~Lin, A.~Maddukuri, S.~Nasiriany, and Y.~Zhu, ``robosuite: A modular simulation framework and benchmark for robot learning,'' \emph{arXiv preprint arXiv:2009.12293}, 2020.

\bibitem{openpi2024}
{Physical Intelligence}, ``openpi: Open-source robot learning framework for the $\pi_0$ and $\pi_{0.5}$ models,'' \url{https://github.com/Physical-Intelligence/openpi}, 2024, accessed 2026-08.

\end{thebibliography}

\end{document}